\documentclass[sigconf]{acmart}

\usepackage{booktabs} 
\usepackage[linesnumbered,lined,commentsnumbered, ruled]{algorithm2e}
\usepackage{multirow}

\setcopyright{acmcopyright}

\acmConference[SAC'20]{ACM SAC Conference}{March 30-April 3, 2020}{Brno, Czech Republic} 
\acmYear{2020}
\copyrightyear{2020}

\acmArticle{4}
\acmPrice{15.00}

\setcopyright{rightsretained}

\begin{document}
\title{Template co-updating in multi-modal human activity recognition systems}

\author{Annalisa Franco}
\orcid{0000-0002-6625-6442}
\affiliation{%
  \institution{University of Bologna}
  \streetaddress{Via dell'Università, 50}
  \city{Cesena} 
  \state{Italy} 
  \postcode{47522}
}
\email{annalisa.franco@unibo.it}

\author{Antonio Magnani}
\orcid{}
\affiliation{%
  \institution{University of Bologna}
  \streetaddress{Via dell'Università, 50}
  \city{Cesena} 
  \state{Italy} 
  \postcode{47522}
}
\email{antonio.magnani@unibo.it}

\author{Dario Maio}
\orcid{0000-0002-0094-0022}
\affiliation{%
  \institution{University of Bologna}
  \streetaddress{Via dell'Università, 50}
  \city{Cesena} 
  \state{Italy} 
  \postcode{47522}
}
\email{dario.maio@unibo.it}

\renewcommand{\shortauthors}{A. Franco et al.}

\begin{abstract}
Multi-modal systems are quite common in the context of human activity recognition; widely
used RGB-D sensors (Kinect is the most prominent example) give access to parallel data
streams, typically RGB images, depth data, skeleton information. The richness of multimodal
information has been largely exploited in many works in the literature, while an
analysis of their effectiveness for incremental template updating has not been investigated so
far. This paper is aimed at defining a general framework for unsupervised template updating
in multi-modal systems, where the different data sources can provide complementary
information, increasing the effectiveness of the updating procedure and reducing at the same
time the probability of incorrect template modifications.
\end{abstract}

%
%

\copyrightyear{2020} 
\acmYear{2020} 
\acmConference[SAC '20]{SAC '20: The 35th ACM/SIGAPP Symposium on Applied Computing}{March 30-April 3, 2020}{Brno, Czech Republic} 
\acmBooktitle{SAC '20: The 35th ACM/SIGAPP Symposium on Applied Computing (SAC '20), March 30-April 3, 2020, Brno, Czech Republic}
\acmDOI{10.1145/3341105.3374085}
\acmISBN{978-1-4503-6866-7/20/03}


\maketitle

\section{Introduction}
The continuous advances in sensing technologies and networking infrastructures enable the development of intelligent systems able to provide real-time analysis of specific situations of interest in a smart home environment, with the aim of enhancing the quality of life of the occupants. For instance, in the field of healthcare, particular attention is generally devoted to systems able to detect and recognize dangerous situations and to provide prompt alarms. In this context a number of interesting contributions have been recently proposed for human activity recognition, based either on the use of specific sensors for data acquisition or on the use of computer vision techniques able to infer human behaviour from images. We will focus in this work on vision-based techniques, whose adoption enhances - from our point of view - the unobtrusiveness of the monitoring system since they do not require users to wear sensors of any kind. \newline
In the specific context of the home environment, the acquisition of a large amount of training data is quite difficult and unlikely. The home environment is usually characterized by a very limited number of users, and also most of the reference benchmarks for activity recognition reproduce a “small-size” scenario, with few users and few activity samples per user. We are confident that in this scenario “traditional” computer vision techniques can achieve good results and real time processing capabilities even with limited computational power, while techniques based on deep learning are more difficult to apply. On the other hand, if we think at an home environment where the hypothetical monitoring system is continuously checking the ambience to detect possible anomalies and/or to understand human actions, it's clear that a huge amount of unlabelled data can be easily collected. On the contrary, labelled training data are often scarce, with a few video samples for each activity to be recognized; this is in our opinion the realistic scenario where huge amount of training data is very unlikely to be available. We believe therefore that the implementation of incremental updating techniques is mandatory in advanced recognition systems to fully exploit the richness of data that the specific scenario naturally provides. Moreover, many works in the literature are multi-modal in nature, in the sense that they combine different kinds of input data (e.g. RGB images, skeleton, depth data) jointly acquired by some well-known sensors such as Kinect\textsuperscript{\textregistered}. The template updating procedure could thus rely in many cases on different data sources whose combined use can in principle improve the effectiveness of the updating procedure, reducing at the same time the probability of selecting wrong data that could deteriorate the initial templates. Finding a good trade off between the need of adding new information to the initial templates and avoiding updating errors that could compromise them is, in fact, the main challenge in this problem.\newline
This paper proposes an incremental co-updating technique, based on the joint analysis of RGB images and human skeleton information acquired with the Kinect\textsuperscript{\textregistered} sensor. The proposed approach is semi-supervised, i.e. we suppose to have a small initial training set for the creation of the base templates which are subsequently updated in a totally unsupervised way. To the best of our knowledge, only a few works in the literature propose template updating techniques for human activity recognition (see next section) based exclusively on RGB images while the possibility of co-updating based on multiple data sources has not been investigated so far. \newline
The rest of the work is organized as follows: in section \ref{relatedWorks} a discussion of the template updating techniques proposed for human activity recognition is reported, section \ref{proposedApproach} describes the proposed co-updating approach for a generic multi-modal system and presents a specific implementation based on RGB images and skeleton data, the experimental results are reported and discussed in section  \ref{experimentalResults} and finally section \ref{conclusions} draws some concluding remarks and provides ideas for future research.

\section{Related works}
\label{relatedWorks}
A general review of the huge literature on human activity recognition goes beyond the scope of this work. Interested readers can refer to \cite{2018arXiv180611230K} and \cite{articleSurvey2019} for good recent surveys. Most of the existing approaches, however, focus on static activity models, where all the training samples are supposed to be labeled and available at the time the model is first computed.\newline Only a few works address the problem of dynamic template updating. In \cite{5459374} an incremental approach based on a feature tree is proposed; the feature tree grows as new data become available, but this requires maintaining all the training and updating samples and this aspect limits its practical applicability. The authors of \cite{6095611} describe an updating technique based on human tracking in video sequences. In this case a manual annotation of the human body is needed at the beginning of the action clip, so all the updating process is supervised to some extent and unfeasible for our purposes. An active learning technique based on the idea of adding new weak classifiers for new incoming instances is presented in \cite{6909502}. The whole method is based on STIP features \cite{Laptev2005} extracted from RGB images. De Rosa et al. propose in \cite{DEROSA201748} a general framework for active incremental recognition of human activities where the feature space used to represent information is gradually covered with balls centered on samples selected from the stream. Finally, the authors of \cite{7244231} propose a framework for continuous learning based on deep hybrid feature models. In particular, the approach is aimed at automatically learning the optimal feature models for activity recognition exploiting a deep autoencoder, and at continuously updating the templates; to this last purpose, a selection criteria is defined to identify, among the accumulated samples, the best subset for updating. Some of the approaches in the literature are very interesting and exhibit promising performance however, to the best of our knowledge, all of them focus on a single data source (RGB images in most cases).\newline
Multiple data sources have been successfully exploited in other contexts, in particular for multi-modal biometric systems \cite{RoliCoUpdating}; the idea is that systems based on different characteristics provide complementary performance, since each recognizer is expected to assign correct labels to certain input data which are difficult for the other and vice-versa. We will explore the applicability of this principle to the problem of activity recognition where different data sources (e.g. RGB, skeleton, depth data), possibly independent, are likely to be available when common acquisition devices such as Kinect\textsuperscript{\textregistered} are used.  

\section{Proposed approach}
\label{proposedApproach}
The aim of this work is to propose a general framework for template co-updating based on the analysis of multiple data sources. The algorithm will be described in the next section without any assumption about the specific features used; then a possible implementation based on the combination of information from RGB images and human skeleton will be described and used for the experimental validation of our proposal.
\subsection{The general template co-updating algorithm}
The template co-updating procedure exploits $n$ types of information derived from different data sources. For each source, a specific classifier $Cl^k, k=1,..,n$ is pre-trained on a set of $a$ activities; the resulting  templates are: $T^k = \{T_1^k,...,T_a^k\}$. 
The basic idea of our co-updating algorithm is that when the prediction of an input sequence operated by a specific classifiers (at least one) fulfills a series of reliability criteria, then that sequence will be used to update all classifiers (of course each with the specific data source). This way we think that it will be possible to accept for the other classifiers also data rather far from the existing templates thus increasing their representativeness. 
With the aim of determining the robustness and reliability of a prediction, we considered the following criteria. 
\subsubsection{Reliability of each classifier}
A classifier may exhibit poor reliability in predicting specific classes of activities, and provide very good results on others. The reasons may be different, including the possible and repeated occlusion of body parts or the poor representativeness of the templates used. This can obviously lead to rapid degeneration of the quality of predictions concerning a specific classifier. Such a phenomenon becomes critical in a co-updating context, where, if not adequately addressed, one classifier may corrupt the templates of others. Based on the reliability of the various data sources in relation to the initial templates, our approach assigns to each classifier an activity-specif weight, proportional to its ability to correctly identify activity sequences of each class. To pursue this goal, we determine, for each classifier $Cl^k$, a set of weights  $w_{i}^k = \{w_1^k,...,w_a^k\}$ for the different activities based on the classifier \textit{precision} (as defined in Section \ref{experimentalResults}). Among the others, we adopted this metric because it minimizes the likelihood of accepting false positives for a specific class, a fundamental characteristic in this scenario, where accepting a false positive is certainly the most critical type of error.

\subsubsection{Degree of certainty of a prediction}
The decision of exploiting an incoming sequence $s_j$ for template updating relies on the robustness of its classification by the different classifiers. To this end, let's suppose that the classifier $Cl^k$ produces a distribution of probabilities $\textbf{p}^k=[p^k_1,..,p^k_a]$ for the sequence $s_j$ over the $a$ activity classes, and let's suppose that the two most probable classes are, respectively, $c^k_1$ and $c^k_2$. We define the \textit{degree of certainty} of the prediction ($s_j$ belongs to class $c^k_1$) such as:
\[doc({c^k_1})=1-\frac{\textbf{p}^k[c^k_2]}{\textbf{p}^k[c^k_1]}\]
where $p^k[{c^k_2}]$ and $p^k[{c^k_1}]$ are the two highest probability estimates offered by the classifier. 
The rationale behind this choice is that when the two highest probabilities are both high and similar, the prediction is very uncertain. Such a simple metric can be particularly useful in cases of strong similarities between classes (\textit{e.g., drink/talking on phone}) and allows to exclude potential risky updates that could affect the robustness of the approach.
The degree of certainty of the prediction is then weighed for the previously determined set of weights and define the \textit{credibility} as:
\[cre({c^k_1}) = doc({c^k_1}) * w_{c^k_1}^k\]
\subsubsection{Restrictive sequence acceptance rules}
In order to define the criteria of acceptance of a new sequence, several aspects need to be taken into account. The first distinction is whether or not all classifiers agree on the predicted class (line 8 Algorithm \ref{algo-recog}). In the first case, to avoid a common misclassification, the algorithm exploits two different parameters of acceptability: $\delta_{cre}$ defines a threshold for the credibility (0.35 in our experiments) while $\delta_{close}$ defines a closeness threshold (0.2). On the one hand, $\delta_{cre}$ certifies that all predictions are considered robust, on the other, $\delta_{close}$ defines a common closeness of the consensus. Indeed, a classifier may have a degree of credibility higher than the relative threshold, but distant from the others; this implies a partial dissent in the common choice. Of course, both the thresholds could be weighed by the number of classifiers, relaxing these constraints. 
On the other hand, if classifiers do not agree, the prediction with the highest credibility (which must necessarily be higher than $\delta_{cre}$) will be considered (line 9 Algorithm \ref{algo-recog}). In fact, the degree of credibility must be sufficiently higher than the one presented by others. For this purpose, $\delta_{diff}$, which parameterises the supremacy of one prediction over the others, is exploited. For our experiments we fixed this parameter to 0.2. These constraints are defined in Algorithm \ref{algo-recog}. If the algorithm is able to assign a reliable class label to the new sequence observed, according to the rules described above, the new sample will be included in the buffer of labeled samples $B$ and used for template updating (for all the classifiers). Otherwise, the sample will be queued to the buffers of unlabeled samples $U$. After each incremental update of the classifier, our framework will attempt to re-assign a label to the queued unlabeled samples (line 17 Algorithm \ref{algo_coup}); after the updating, in fact, they could be better recognized.
\subsubsection{Template preservation}

Finally, it is realistic to assume that buffers are limited, so that when they reach their maximum capacity, some samples have to be removed. The strategy adopted is to preserve the most expressive samples for a given class. Therefore, even if it may seem counterintuitive, the underlying assumption is to search for those buffered samples for which the classifiers show the highest uncertainty. Indeed, if the templates with the highest degree of confidence were preserved, it would be legitimate to assume that the classifier is reducing the expressiveness of a specific activity representation; the most useful templates could potentially be excluded from an adequate representation of intra-class variations. Clearly, the search policy is based on the class of the element that is causing the buffer overflow. 
In addition, to avoid over-representation of a specific class, the places in the buffer are evenly distributed among the different classes; therefore, adding a template is bound to an additional parameter that defines the maximum number of buffer elements for a given class. The maximun buffer size has been fixed in our tests to 170, and the buffer is initialized with the samples of the training set, in order to avoid any forgetting effect. 

\IncMargin{1em}
\begin{algorithm}

\SetKwData{Left}{left}
\SetKwData{F}{F}
\SetKwData{W}{W}
\SetKwData{Ylbl}{y}
\SetKwData{labelSet}{$Y^{f}$}
\SetKwData{bufferMax}{bufferMax}
\SetKwData{UnlabeledSet}{$U^{f}$}
\SetKwData{Unlabeled}{$U$}
\SetKwData{Buffer}{$B^{f}$}
\SetKwData{Classifier}{$Cl^{f}$}
\SetKwData{FeatureChannel}{$f$}
\SetKwData{flagTemplate}{$newTemplate$}
\SetKwData{Up}{up}

\SetKwFunction{Union}{Union}
\SetKwFunction{UpdateBuffer}{UpdateBuffer}
\SetKwFunction{UpdateClassifier}{UpdateClassifier}
\SetKwFunction{ExtractFeatures}{ExtractFeatures}
\SetKwFunction{RecognizeTemplate}{AssignClassLabel}
\SetKwFunction{FindCompress}{FindCompress}
\BlankLine
\emph{Initialize variables}\;
\ForEach{new sequence $s_j$}{
    \F$\leftarrow$ \ExtractFeatures{$s_j$}\; 
    \Ylbl$\leftarrow$ \RecognizeTemplate{\F}\;
    \eIf(\tcp*[f]{sequence not labeled}){\Ylbl$=-1$}{ 
    $\UnlabeledSet \leftarrow \UnlabeledSet \cup  \FeatureChannel \; \forall \FeatureChannel \in F$ \;
    }(\tcp*[f]{sequence labeled})
    {
    $\Buffer \leftarrow \Buffer \cup \FeatureChannel \; \forall \FeatureChannel \in F$ \; 
    $\labelSet \leftarrow \Buffer \cup \Ylbl \; \forall \FeatureChannel \in F$ \; 
    $\flagTemplate \leftarrow 1 \;$
    }
    \If{$\Buffer \geq \bufferMax $}{
        $\Buffer \leftarrow \UpdateBuffer{\Buffer,\labelSet} \; \forall \FeatureChannel \in F$ \;
    }
    
    \If(\tcp*[f]{classifiers update}){$\flagTemplate = 1$}{
        $\Classifier \leftarrow \UpdateClassifier{\Buffer,\labelSet} \forall \FeatureChannel \in F$ \; 
        Retry recognizing all unlabeled elements $\in U$ and update classifiers if needed;\newline
         $\flagTemplate \leftarrow 0 \;$
    }
}
\caption{Template Co-Updating}\label{algo_coup}
\end{algorithm}\DecMargin{1em}

\IncMargin{1em}
\begin{algorithm}
\SetKwData{Prediction}{$\textbf{p}^k$}
\SetKwData{weight}{$w_{c^k_1}^k$}
\SetKwData{F}{F}

\SetKwData{doc}{$doc(c^k_1)$}
\SetKwData{lblPredicted}{$c^k_1$, $c^k_2$}
\SetKwData{CommentOrdering}{Sort \Prediction by confidence}
\SetKwData{ThrCloseness}{$\delta_{close}$}
\SetKwData{ThrDiff}{$\delta_{diff}$}
\SetKwData{ThrRel}{$\delta_{cre}$}
\SetKwData{rel}{$cre(c^k_1)$}
\SetKwFunction{ArgMax}{MostProbableClasses}
\SetKwFunction{ProbabilistiPrediction}{ProbabilisticPrediction}
\SetKwInOut{Input}{input}\SetKwInOut{Output}{output}
\Input{\F=\{$f^1,...,f^n$\}}
\Output{A predicted class or $-1$}

\BlankLine
\emph{Initialize variables}\;
\ForEach{feature channel $f^k$ in $F$}{
    $\Prediction \leftarrow \ProbabilistiPrediction {$f^k$} $\;
    $\lblPredicted \leftarrow \ArgMax{\Prediction}$ \;
    $\doc \leftarrow 1 - \textbf{p}^k[c^k_2] \setminus \textbf{p}^k[c^k_1]$\;
    $\rel \leftarrow \doc \ast \weight $\;
}
\lIf{$(c^i_1 = c^j_1 = c \medspace \forall (Cl^i, Cl^j)) \land ( cre(c^i_1)\geq \ThrRel \medspace \forall i) \land \newline (|cre(c^i_1) - cre(c^j_1)| < \ThrCloseness \forall \medspace (Cl^i, Cl^j))$\newline }{
    \Return $c$ 
}

\eIf{$\exists\medspace Cl^k | \medspace (\rel \geq \ThrRel) \land\newline (|\rel - cre(c^i_1)| \geq \ThrDiff \medspace \forall i)$}{
    \Return $c^k_1$
}{
\Return $-1$ \;
}
\caption{AssignClassLabel(F)}\label{algo-recog}
\end{algorithm}\DecMargin{1em}

\subsection{An implementation based on RGB and skeleton data}

\subsubsection{RGB data representation}
In order to represent RGB information, among the possible alternatives we adopted Improved Dense Trajectories \cite{Wang2013}, well-known for their excellent performance in action recognition tasks. These trajectories are a composition of different features extracted from each video sequence, specifically: Histogram of Oriented Gradient (HOG), Histogram of Optical Flow (HOF) and Motion Boundary Histogram (MBH).
The features have been extracted from each video using the code published on the INRIA website \footnote{\href{http://lear.inrialpes.fr/people/wang/improved\_trajectories}{http://lear.inrialpes.fr/people/wang/improved\_trajectories}}. Similar to \cite{DEROSA201748}, we kept the default parameters and only reduced the length of the trajectory (from 15 to 8 frames).
The extracted trajectories are accumulated and each of the four features used (HOG, HOF, MBHx and MBHy) has been encoded using a Bag of Word model (BoW) \cite{wang:inria-00439769} with K=500. Therefore, each video is described by a histogram of 2000 elements obtained by concatenating the individual descriptors.
The classifier adopted is a linear incremental SVM with stochastic gradient descent (SGD) learning\footnote{\href{https://scikit-learn.org/stable/modules/generated/sklearn.linear\_model.SGDClassifier.html}{Scikit-Learn SGD Classifier}}.

\subsubsection{Skeleton data representation}
We recently proposed in \cite{DBLP:conf/iciap/FrancoMM17} an activity recognition approach based on skeleton joint orientations. A posture representation has been defined based exclusively on angle information, derived from both the joint position and orientation. Angle features derived from skeletons have the great advantage of being intrinsically normalized and independent from the user's physical structure. They also guarantee a good degree of invariance with respect to pose and view changes since all the angles are computed with respect to the subject's coordinate system.\\ 
Each frame of a video sequence is represented by a set of angles derived from the human skeleton, which summarize the relative positions of the different body parts. 
The Kinect\textsuperscript{\textregistered} SDK represents the human skeleton as a set of $d$ joints $J=\left\lbrace j_1, j_2,...,j_d\right\rbrace$; each joint $j_i=\left(\mathbf{p_i}, \overrightarrow{\mathbf{o_i}}\right)$ is described by its 3D position $\mathbf{p_i}$ and its orientation $\overrightarrow{\mathbf{o_i}}$ with respect to ``the world''. To encode the user posture, we defined three types of angles:
\begin{itemize}
	\item $\theta_{ab}$: angle between the orientations $\overrightarrow{\mathbf{o}_a}$ and $\overrightarrow{\mathbf{o}_b}$ of joints $j_a$ and $j_b$. $\theta_{ab}$ angles are computed from a set of $m$ ($m=8$) couples of joints $A_\theta$. 
	\item $\varphi_{ab}$: angle between the orientation $\overrightarrow{\mathbf{o}_a}$ of $j_a$ and the segment $\overrightarrow{j_aj_b}$ connecting $j_a$ to $j_b$ (we can consider the segment as the bone that interconnects the two joints). $\varphi_{ab}$ angles are computed from a set of $n$ couples ($n=16$) of joints $A_\varphi$. 
	\item $\alpha_{bac}$: angle between the segment $\overrightarrow{j_aj_b}$ connecting $j_a$ to $j_b$ and $\overrightarrow{j_aj_c}$ that connects $j_a$ to $j_c$. $\alpha_{bac}$ angles are computed from a set of $s$ triplets ($s=4$) of joints $A_\alpha$.
\end{itemize}

The skeleton estimation provided by Kinect\textsuperscript{\textregistered} is quite reliable for the joints of the upper part of the body, which contains most of the information needed for activity recognition, while the legs are generally quite unreliable, often occluded and almost static . For this reason only a subset of the possible angles is considered, mainly obtained from the joints of the upper part of the body. All the details about the selected angles ($A_\theta$, $A_\varphi$ and $A_\alpha$) can be found in \cite{DBLP:conf/iciap/FrancoMM17}.\\
Each skeleton $SK_t$ of the video sequence is represented by a vector obtained as the ordered concatenation of the values of $\theta_i\ |\ i\in A_\theta$, $\varphi_j\ |\ j\in A_\varphi$, $\alpha_k\ |\ k\in A_\alpha$ \[\mathbf{v}_i=\left(\theta_1,...,\theta_m, \varphi_1,...\varphi_n, \alpha_1,...,\alpha_s\right)\] of size ($m+n+s$) where $m=|A_\theta|$, $n=|A\varphi|$ and $s=|A_\alpha|$. \\
The complete video sequence $S$ is finally encoded using a BoW model where each activity is represented as an histogram of occurrences of some reference postures.
The same classifier exploited for RGB information has been used here as well.
\section{Experiments}
\label{experimentalResults}
The proposed approach has been validated with extensive experiments where the proposed co-updating technique is compared with a batch template creation and a fully supervised incremental updating.

\subsection{Database and protocol}

\begin{figure*}[!ht]
  \includegraphics[width=0.95\textwidth]{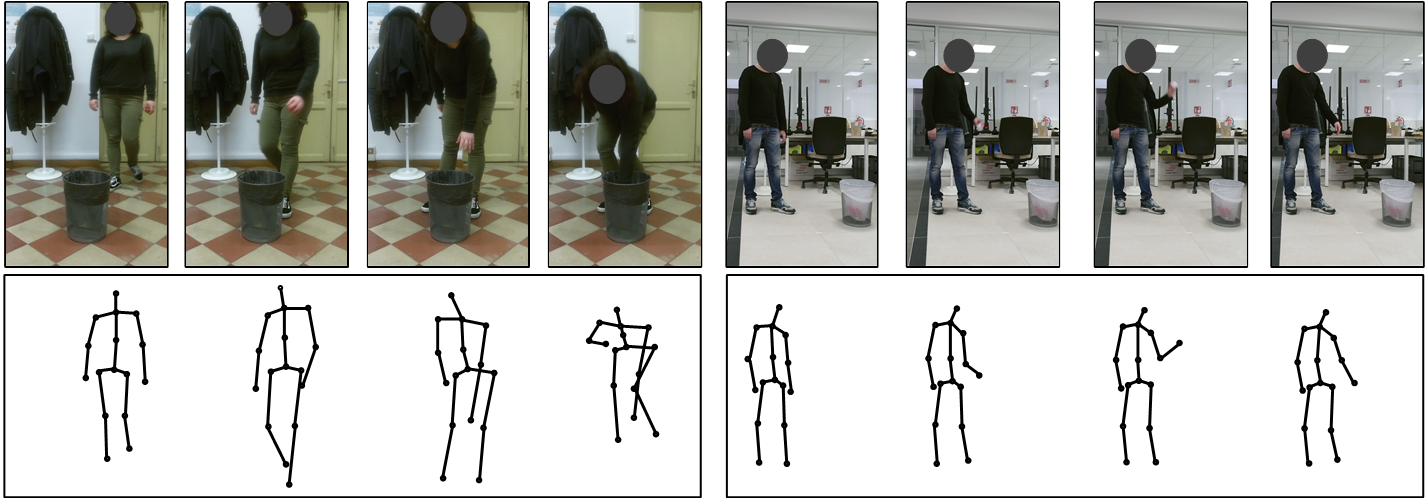}
  \caption{Sample frames (RGB and skeleton) for the "throw something in bin" activity.}
  \label{fig:bin1}
\end{figure*}

The independence of the data sources used for co-updating is a key factor for the effectiveness of the process. Most of the huge databases for activity recognition only include RGB images and are therefore unfeasible for our application. Only a couple of datasets, the Cornell Activity Datasets CAD-60 and CAD-120, include the two modalities (RGB and skeleton with joint orientation information) that we considered for our practical implementation. The size of the two CAD datasets is unfortunately very limited, with data taken from only 4 sujects, and we think that a validation on this data would not be so meaningful. For this reason we performed all the experiments on the Office Activity dataset (OAD), acquired in our laboratory and publicly available \footnote{\href{http://smartcity.csr.unibo.it/activity-recognition/}{Smart City Lab}}. It includes a total of 560 video sequences of 14 activities (\textit{drinking, getting up, grabbing an object from a shelf, pour a drink, scrolling book pages, sitting, stacking items, take objects from a shelf, talking on the phone, throwing something in the bin, waving hand, wearing coat, working on computer, writing on paper}); each activity is performed twice by 20 subjects in a different environment from several perspectives based on the activity being performed. The execution of the different activities was loosely supervised; a generic definition of the activity was given to each subject without any precise indication about how it should be carried out, and this produces a significant intra-class variability. We are aware that number of subjects in the database is still not high, but we believe that the tests reported in this paper can give a clear idea of the potentialities of this algorithm. 
The skeletal data provided are composed of the 3D positions of 25 tracked joints and the orientations of 19 of them. RGB and depth images will be released with permission and in accordance with the General Data Protection Regulation (GDPR, EU no. 2016/679).\newline

For testing the proposed updating technique, we followed the common "new person" protocol, meaning that disjoint subjects are used for training, updating and testing. In particular, the available subjects are randomly partitioned as follows:\textit{training set} ($TR$) 20\% (4 subjects, 112 sequences), \textit{updating set} ($UPD$)  50\% (10 subjects, 280 sequences), \textit{testing set} ($TE$) 30\% (6 subjects, 168 sequences). The experiments are repeated seven times with different subject partitions and the average results are finally computed. The performance is reported in the form of overall recognition accuracy (percentage of activities correctly classified) and confusion matrix, where the rows correspond to the real activity label and the columns to the estimated one. Moreover, we report precision \textit{P} and recall \textit{R} values, computed as follows: \[P=\frac{TP}{TP+FP},  R=\frac{TP}{TP+FN}\] where \textit{TP, FP} and \textit{FN} represent respectively the True Positives, False Positives and False Negatives which can be easily derived from the extra-diagonal elements of the confusion matrix.
For evaluation purposes, the unsupervised co-updating approach is compared to two supervised approaches. In particular, the performance are reported for:
\begin{itemize}
    \item \textit{Proposed co-updating} (\textbf{Co-Updating}): the initial templates are created from the subjects in the train set $TR$ and subsequently incrementally updated with the set $U$ based on the approach proposed in this work (see section \ref{proposedApproach}).
    \item  \textit{Supervised template updating} (\textbf{Supervised Updating}): the initial templates are created from the subjects in $TS$, and then are updated sequentially using the subjects in $UPD$ exploiting the real activity labels; all the incoming information is used in this case correctly, so this approach gives an upper bound to the performance that can be achieved with an incremental learning strategy.
    \item \textit{Batch template creation} (\textbf{Batch}): all the subjects of training and updating sets ($TR \cup UPD$) are exploited for the initial template creation; this approach gives an idea of the top-performance that could be obtained if a huge amount of training data were available at the time of initial template creation. The system is static, no updating is carried out in this case.
\end{itemize}
In all cases, the performance are measured on the testing subjects in $TS$.
\subsection{Results}
Table \ref{tab:results_res} summarizes the results in terms of precision and recall measured for the proposed approach, as well as for the two supervised techniques taken as reference systems. Of course the supervised approaches provide an upper bound to the performance that can be achieved; in particular the batch approach represents the most favorable case where a huge amount of data is available from the beginning. The performance indicators are given for the separate modalities (RGB and skeleton) as well as for their fusion. The results measured with the initial template are compared to those obtained after updating to appreciate the effects of template updating.\newline

\begin{figure*}[!htb]
	\centering
	\includegraphics[width=0.7\textwidth]{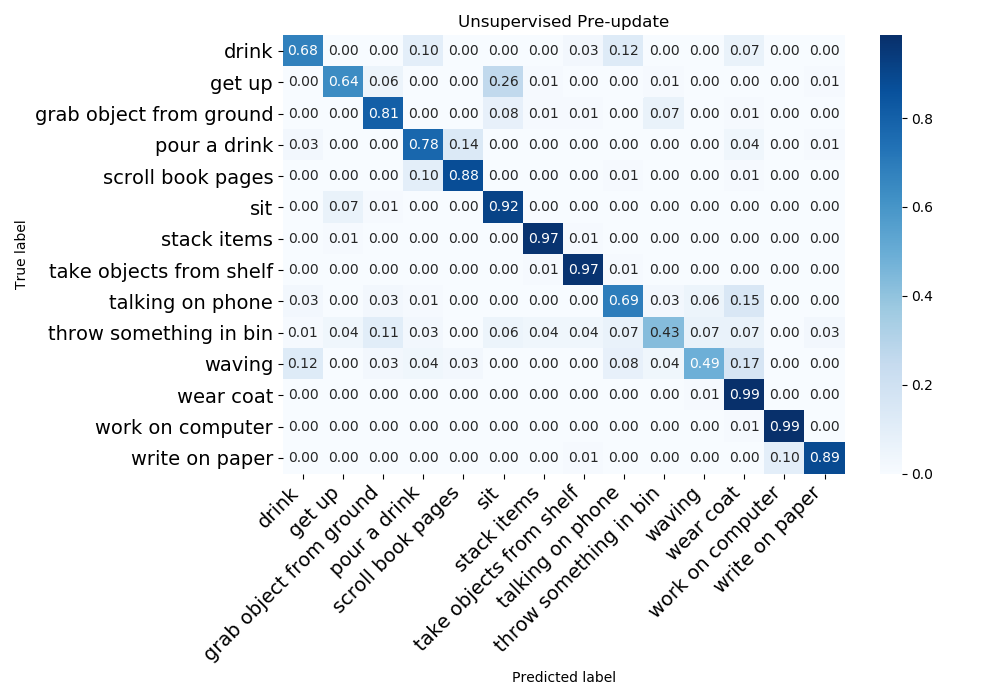}
	\hfill 
	\caption{Confusion matrix using only the training set (i.e., before the application of the template co-updating algorithm).}
	\label{fig:cm_preupd}
\end{figure*}  

\begin{figure}[!ht]
	\centering
	\includegraphics[width=0.48\textwidth]{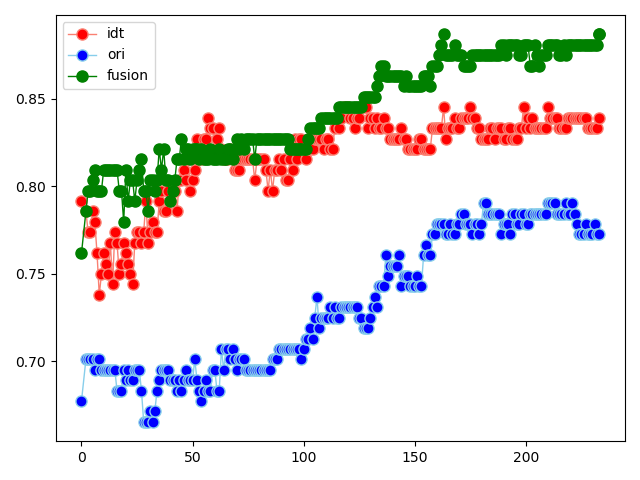}
	\hfill 
	\caption{Activity recognition accuracy trend during unsupervised template co-updating (as a function of the number of updates performed). The three curves represent the RGB templates (red), the skeleton templates (blue) and their fusion (green).}
	\label{fig:example1}
\end{figure}

RGB features seem to be generally more stable and reliable than skeleton data, even if the difference is overall quite limited. The initial confusion matrices of the two modalities confirm that they are quite complementary. Such independence is very useful during co-updating; in fact, new incoming samples quite far from the existing template for a given feature can often be accepted due to the high confidence of the other; this is indeed the ultimate objective of template updating: adding new, different samples to the template to increase its representativeness. The accuracy trend during the different updating steps in one of the different runs of experiments is given in Figure \ref{fig:example1}. The trend is positive for the single modalities and, as expected, also for their fusion. The skeleton templates greatly benefit from the updating procedure, thanks to the support of RGB templates. \newline

\begin{figure*}[!htb]
	\centering
	\includegraphics[width=0.7\textwidth]{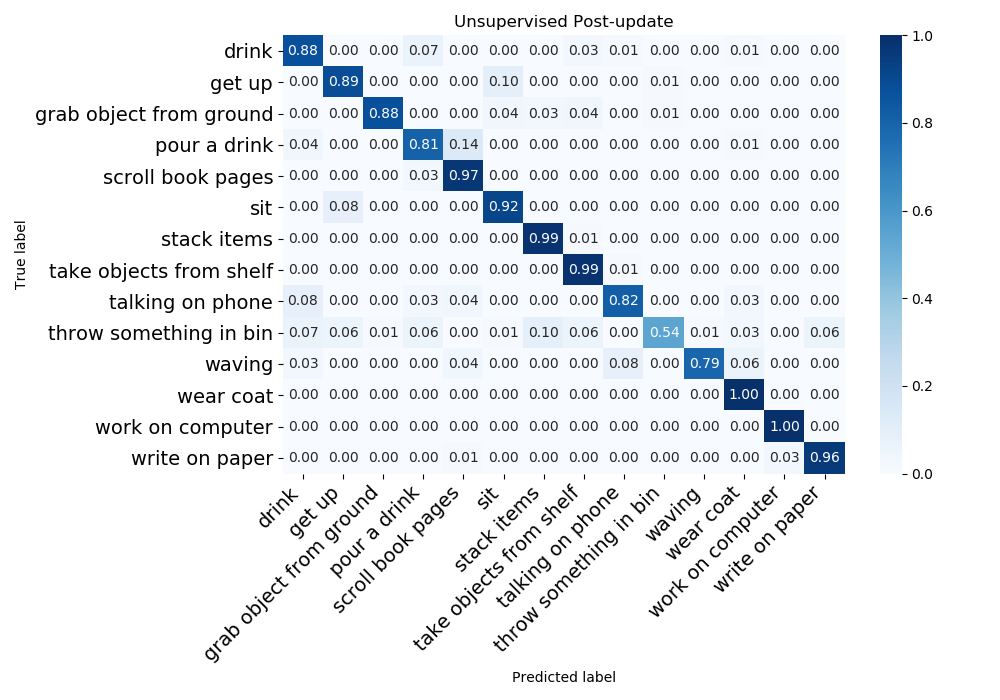}
	\hfill 
	\caption{Confusion matrix after the application of the template co-updating algorithm.}
	\label{fig:cm_postupd}
\end{figure*}

When the performance of the unsupervised approach are compared to the batch of the supervised one, we can observe very similar values of precision and recall. This very positive result confirms that the unsupervised approach exploits at best the updating set, i.e. accepts a high number of unknown sequences and correctly uses them for template updating. \newline
The effects of our co-updating procedure can be better analyzed looking at the confusion matrices of Figure \ref{fig:cm_preupd} and Figure \ref{fig:cm_postupd} referred, respectively, to the initial templates and to the final templates obtained with unsupervised updating. Some activities were already recognized with a good level of accuracy even with the initial templates (e.g. stack items, take objects from shelf, wear coat, work on computer) and such performance is preserved by the updating procedure; we can conclude that wrong updates are very rare, totally absent for most activities, and the initial templates are not corrupted. For other activities, probably characterized by a higher degree of variability between different subjects, the initial templates provide poor performance; template updating greatly improves the results in several cases (e.g. drink, get up, waving or talking on phone). Particularly interesting is the improvement observed for "get up", very similar from the point of view of body posture to "sit". Several of the surviving errors are quite comprehensible if we consider that the mistaken activities often share common body positions and movements. A very particular case is represented by the "throw something in bin" activity which has been interpreted by the volunteers with great fantasy, in very different fashions (see Figure \ref{fig:bin1} for two samples of this activity). This is the reason why the performance increment is limited in this case and the final accuracy is still sub-optimal.    

\begin{table}[]
\begin{tabular}{|c|c|c|c|c|c|}
\hline
\multicolumn{2}{|c|}{\multirow{2}{*}{}} & \multicolumn{2}{c|}{\textbf{Initial templates}} & \multicolumn{2}{c|}{\textbf{Final templates}} \\ \cline{3-6} 
\multicolumn{2}{|c|}{} & \textit{Precision} & \textit{Recall} & \textit{Precision} & \textit{Recall} \\ \hline
\multirow{3}{*}{\textbf{Skeleton}} & Co-Upd. & 0.701 & 0.697 & 0.758 & 0.749  \\ \cline{2-6} 
 & Sup. Upd. & 0.701 & 0.697 & 0.767 & 0.752 \\ \cline{2-6} 
 & Batch & 0.785 & 0.776 & // & // \\ \hline
\multirow{3}{*}{\textbf{RGB}} & Co-Upd. & 0.798 & 0.788 & 0.887 & 0.872 \\ \cline{2-6} 
 & Sup. Upd. & 0.798 & 0.788 & 0.907 & 0.891 \\ \cline{2-6} 
 & Batch & 0.925 & 0.917 & // & // \\ \hline
\multirow{3}{*}{\textbf{Fusion}} & Co-Upd. & 0.799 & 0.794 & 0.893 & 0.887 \\ \cline{2-6} 
 & Sup. Upd. & 0.799 & 0.794 & 0.923 & 0.922 \\ \cline{2-6} 
 & Batch & 0.944 & 0.943 & // & // \\ \hline
\end{tabular}
\caption{Comparison between the proposed co-updating procedure, the supervised updating (where real activity labels are exploited) and the batch updating (where all the training samples are available for initial template creation)}.
\label{tab:results_res}
\end{table}

\section{Conclusions}
\label{conclusions}
In this paper a general framework for template co-updating in multi-modal activity recognition systems has been proposed. The validity of the proposal has been assessed with a specific implementation based on RGB and skeleton data extracted from video sequences acquired with the Kinect\textsuperscript{\textregistered} sensor. The results show that jointly exploiting different data modalities allows to greatly improve the initial performance, thanks to the inclusion of new data, previously not adequately represented by the initial templates. \newline
The results obtained are fully satisfactory, however several further improvements are possible: for instance the weights used in the algorithm for the different classifiers are now static while a dynamic updating could better exploit the effects of template updating. The main extension of the approach will be the definition of a strategy for discovering new activity classes and including them in the set of known behaviours.

\bibliographystyle{ACM-Reference-Format}
\bibliography{bibliography} 

\end{document}